# Arabic Handwritten Character Recognition based on Convolution Neural Networks and Support Vector Machine

Mahmoud Shams[1]

Department of Machine Learning and Information Retrieval, Faculty of Artificial Intelligence, Kafrelsheikh University
Kafrelsheikh, Egypt, 33511

Amira. A. Elsonbaty[2]

Department of electronic and Communication
Higher Institute of Engineering and Technology in New Damietta, Egypt.

Wael. Z. ElSawy[3]

Department of Business Information System
Faculty of Management Technology & Information System, PortSaid University, Egypt, 42511

*Abstract*—**Recognition of Arabic characters is essential for natural language processing and computer vision fields. The need to recognize and classify the handwritten Arabic letters and characters are essentially required. In this paper, we present an algorithm for recognizing Arabic letters and characters based on using deep convolution neural networks (DCNN) and support vector machine (SVM). This paper addresses the problem of recognizing the Arabic handwritten characters by determining the similarity between the input templates and the pre-stored templates using both fully connected DCNN and dropout SVM. Furthermore, this paper determines the correct classification rate (CRR) depends on the accuracy of the corrected classified templates, of the recognized handwritten Arabic characters. Moreover, we determine the error classification rate (ECR). The experimental results of this work indicate the ability of the proposed algorithm to recognize, identify, and verify the input handwritten Arabic characters. Furthermore, the proposed system determines similar Arabic characters using a clustering algorithm based on the K-means clustering approach to handle the problem of multi-stroke in Arabic characters. The comparative evaluation is stated and the system accuracy reached 95.07% CRR with 4.93% ECR compared with the state of the art.**

*Keywords—Handwritten Arabic recognition; convolutional neural networks; support vector machine*

## I. INTRODUCTION

In recent times, Arabic letter recognition has become required in many applications especially in this epidemic, as traditional education is being transferred to digital education that may need distance learning via the Internet for children or people who need to learn Arabic. Moreover, it is very difficult to recognize Arabic handwritten language especially to evaluate the correct way for writing Arabic characters for undergraduate or postgraduate students, making scientific research is demanded. Therefore, in this paper, we present an Arabic handwritten character recognition system to be one of the effective keys for Arabic handwritten recognition.

Deep learning techniques are effective tools for feature extraction and classification in computer vision and pattern recognition fields. In this paper, DCNN is proposed to classify and to extract the features of the input images represent the handwritten Arabic characters [1-3].

SVM is an efficient supervised statistical classifier used to classify the input patterns separating the nearest data points using hyperplanes [4-6].

The main contribution of this paper can be summarized as follows:

*1)* Building an architecture based on DCNN to classify and recognize the handwritten Arabic characters using a fully connected layer.

*2)* Using Drop out SVM with the softmax layer to proceed with the input features in the existence of DCNN to boost and compare the result with the fully connected DCNN.

*3)* Handling the problem of multi-stroke similar Arabic handwritten characters.

*4)* A comparison study is performed to evaluate the proposed algorithm with the state of the most recent approaches.

The rest of this paper is organized as follows: Section II illustrates the literature survey of the related work; the proposed algorithm is investigated in Section III; the results and discussion is demonstrated in Section IV; finally, the conclusion and future work are presented in Section V.

## II. RELATED WORK

Recognition of letters and characters are essential in text mining and information retrieval system especially search engine. In the present, there are many attempts to recognize, classify, verify, and identify Arabic handwritten characters using machine learning (ML) approaches.

Arabic handwritten recognition based on SVM and radial basis function (RBF) is presented by Elleuch et al. [7], the error classification rate (ECR) of their proposed algorithm was 11.23%. Moreover in [8] a deep learning approach based on SVM they named as DSVM is composed of three SVMs after using dropout and they achieved ECR 5.86%.

A CNN approach based on the SVM classifier is further utilized by Elleuch [9] to boost the extracted features from Arabic handwritten and the recognition accuracy reached to 97.35% and 93.41% using 24 and 66 classes of HACDB database respectively. Furthermore, a comparative study of the methods depends on CNN and SVM are listed in [10] that indicated that the ECR achieve results 5.83%, and 2.09% using the HACDB database of 66 and 24 classes respectively.

El-Sawy [11] et al presented an algorithm based on CNN for handwritten Arabic digits. They used 10 classes based on fully connected CNN and the accuracy obtained reached 88% with ECR=12%. Moreover in [12], they used CNN for Arabic handwritten character recognition. Their ECR is 5.1% using the ReLU activation function that calculates the maximum values of non-saturated feature maps of the input images. They used fully connected 28 classes CNN with classification accuracy 94.4%.

Younis [13] present DCNN to recognize Arabic handwritten characters and they achieve 94.8% and 97.6% for AIA9k and the AHCD databases, respectively.

Hassan et al. [14] presented an algorithm based on scale-invariant feature transform (SIFT) and SVM for offline handwritten Arabic recognition applied to the AHDB database with a promising recognition rate reached to 99.08%.

For isolated Arabic character recognition, Huque et al. [15] presented a comparative study of using K nearest neighbors (KNN), SVM, and sparse representation classifier (SRC) to recognize Arabic handwritten characters base on feature fusion.

The major problem in Arabic handwritten recognition is the multi-stroke problem that stated there is a great confusion with similar character stroke i.e. there are some may be difficult to recognize such as "ط" and "ظ" the point "." Is the major difference as presented in [16]. Moreover, a hierarchal clustering of the most common stroke characters is presented by Boudelaa [17] et al. A detailed comparative study of using DCNN in Arabic handwritten character recognition is presented by Ghanim et al [18]. They compared their approach using Alex net DCNN with a recognition rate of 95.6% with a dynamic Bayesian network (DBN) and SVM. Moreover, Mustafa et al [19] a deep learning approach is presented to classify and recognize Arabic names, and the accuracy reached 99.14% in the SUST-ARG-names dataset.

The major limitations of the stated woks are the inability to recognize multi-stroke Arabic characters. As well as clustering the similar characters for unsupervised learning methodologies.

In this paper, we present an algorithm based on using DCNN with a fully connected layer and softmax and dropout the input features to be classified using SVM. We can use a k-means clustering algorithm to handle the multi-stroke Arabic characters recognition.

## III. PROPOSED METHODOLOGY

### A. Deep Convolutional Neural Network

CNN is one of the most common and wide range of tools in computer vision and pattern recognition. In this paper, we present DCNN to extract the features of the input image that represent the Arabic handwritten characters. Furthermore, a fully connected layer consists of 28 classes represent the Arabic characters is applied. We used 3 convolutional layers called C1, C2, C3. The applied input images are normalized to 64×64. Then the mask of CNN is 7×7 is applied to the 64 × 64 images to produce a convoluted layer that is pooling in a maximum manner that is called max-pooling. Every convolution layer has 32 mini-batch based on a mini-batch gradient descent optimizer.

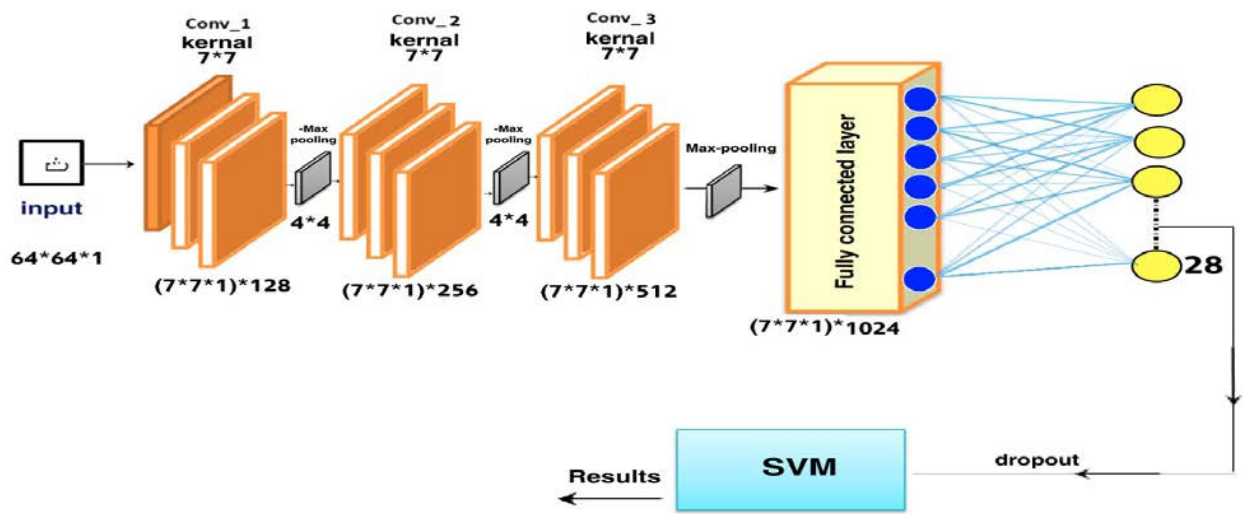

Fig 1. The Proposed DCNN-SVM Architecture.

To ensure the ability of the proposed DCNN to extract more features details, we present three convolutional layers and apply the ReLU activation function. The max-pooling size 4 × 4 is applied after each convoluted layer. Afterward, a fully connected layer that represents 28 classes that are used to classify the input pattern represents the Arabic character.

We present SVM as a statistical kernel classifier after the soft-max drop-out to boost the results of the fully connected layer. Fig. 1 illustrates the proposed DCNN-SVM architecture that investigates all the steps required to recognize and classify the Arabic handwritten characters.

### B. Support Vector Machine

In this paper, to boost the results obtained from DCNN to classify the input patterns with 28 classes which represent the Arabic handwritten characters we present SVM with multi-class labels. SVM can classify the dropout fully connected DCNN to 28 classes that represent all cases of Arabic characters that will be illustrated in detail in the results and discussion section. Table I discuss in detail the architecture of the proposed DCNN-SVM architecture.

This paper proposed an algorithm used to handle the problem of multi-stroke of Arabic handwritten characteristics by using the same protocol investigated in [16]. Moreover, the clustering of similar Arabic characters is performed using the K-means clustering algorithm by quantizing the vectors to 13 clusters that represent the masterstroke and similar characters. The clustering procedure based on k-means is based on the same protocol presented by Ntalianis et al [20]. Table II investigates the 13 clusters that represent Arabic character masterstroke, similar characters, handwritten characters, and English pronunciation for each character.

TABLE I. THE CHARACTERISTICS OF THE PROPOSED DCNN-SVM ARCHITECTURE

| Layer Name. | DCNN-SVM characteristics | |
|---|---|---|
| | *Input Size* | *Kernel Size* |
| Input | 64×64×1 | 7×7 |
| Conv-1 | 64×64×128 | 7×7 |
| Max-Pooling | 4×4 | - |
| Conv-2 | 32×32×256 | 7×7 |
| Max-Pooling | 4×4 | - |
| Conv-3 | 16×16×512 | 7×7 |
| Max-Pooling | 4×4 | - |
| Fully-Connected | 1×1×1024 | - |

TABLE II. THE ARABIC CHARACTERS MASTERSTROKE AND THEIR SIMILAR HANDWRITTEN CHARACTERS

| Arabic Character Master Stroke | Similar Characters | Handwritten Characters | English Pronunciation |
|---|---|---|---|
| ا | ا،أ،إ |  | Alef |
| ب | ب،ت ث،ت ن ي،ى |  | Baa Taa Thaa Noon Yaa |
| ح | ج ح خ |  | Gem Haa Khaa |
| د | د – ذ |  | Dal Zal |
| ر | ر،ز،و |  | Raa Zeen Waw |

| Arabic Character Master Stroke | Similar Characters | Handwritten Characters | English Pronunciation |
|---|---|---|---|
| س | س | | Seen |
| | ش | | Sheen |
| ص | ص | | Saad |
| | ض | | Daad |
| ط | ط | | Taaa |
| | ظ | | Zaaa |
| ع | ع | | Aeen |
| | غ | | Gheen |
| ف | ف | | Faa |
| | ق | | Kaaaf |
| ك | ك | | Kaf |
| | ل | | Lam |
| م | م | | Mem |
| ه | هـ | | Haa |

## IV. RESULTS AND DISCUSSION

In this section, the experimental results were performed in core i7 NVidia 4G-GT 740m GPU environment based on MATLAB 2020 a. The dataset used is founded in [12] with the following characteristics listed in Table III. Furthermore, we applied 840 tested images from 3 persons each person write each character 10 times. A sample of database collected and compared in this paper is shown in Fig. 2.

TABLE III. DATABASE CHARACTERISTICS

| No. of Characters | 16800 |
|---|---|
| No. of participants | 60 |
| No. of writing each character | 10 per participant from "Alef" to "Yaa" |
| No. of the training set (80%) | 13440 (480 images per class) |
| No. of the testing set (20%) | 3360 (120 images per class) |

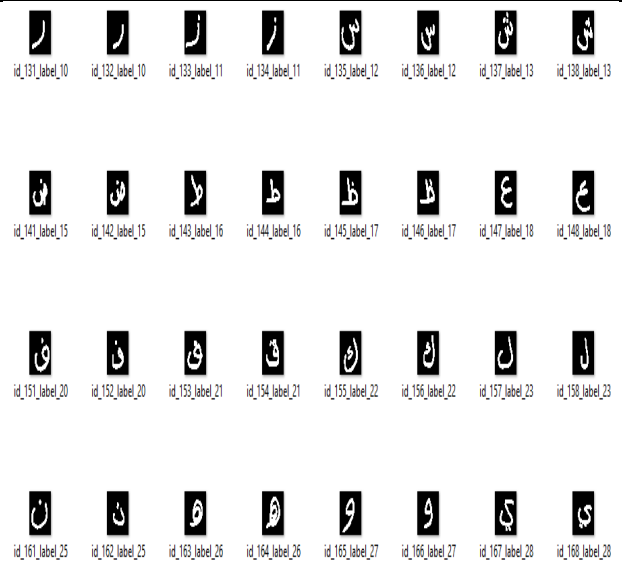

Fig 2. Samples of the Arabic Handwritten Characters [12].

TABLE IV.    THE HYPERPARAMETER VALUES OF THE PROPOSED DCNN-SVM ARCHITECTURE

| Parameter | Value |
|---|---|
| Maximum No. of Iterations | 400 |
| Learning Rate | 0.02 |
| Weight decay | 0.001 |
| Momentum | 0.8 |
| Batch Size | 32 |

The experimental results were executed based on the following fixed hyper-parameter values listed in Table IV as follows:

The above table investigated the maximum number of iteration of the DCNN is 400 with learning rare 0.02 and weighted decay 0.001 and momentum 0.8 with mini-batch gradient descent optimize with batch size 32.

### C. Experiment 1

Based on the hyper-parameter values listed in Table IV, in this experiment we train and test the DCNN-SVM architecture for Arabic handwritten characters using the database in [12] to determine the classification accuracy and the ECR of the proposed system. After 400 iterations the final accuracy of the trained images was 98.08% and 95.07% for the tested images as shown in Fig. 3.

The ECR is determined by as (100 – corrected classified patterns). In our case, the ECR is 1.92% and 4.93% in training and testing, respectively.

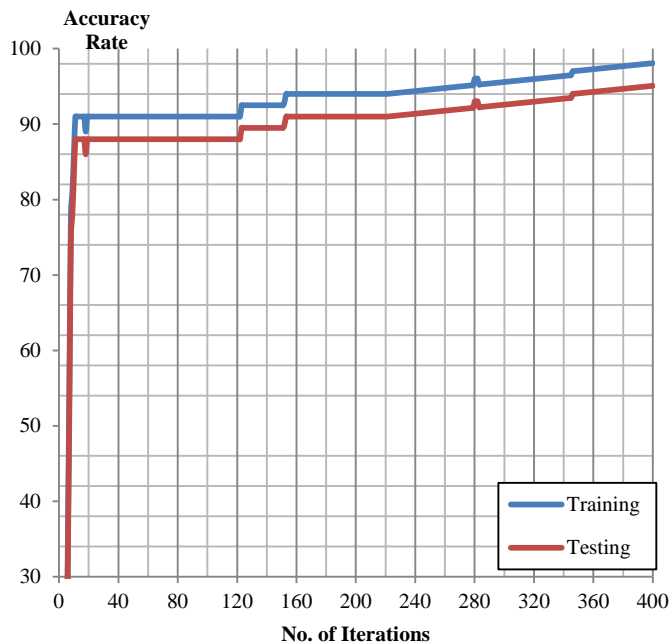

Fig 3.    The Accuracy Rate of the Proposed DCNN-SVM Architecture.

TABLE V.    AVERAGE CORRECT AND ERROR CLASSIFICATION RATE OF THE TESTED HANDWRITTEN ARABIC CHARACTERS

| Handwritten Characters | Correct Classification Rate (%) | Error Classification Rate (%) |
|---|---|---|
| أ | 95.25 | 4.75 |
| ب | 95.21 | 3.79 |
| ج | 97.20 | 2.8 |
| د | 97.23 | 2.77 |
| ر | 93.21 | 6.79 |
| س | 92.54 | 7.46 |
| ص | 94.02 | 4.98 |
| ط | 90.25 | 9.75 |
| ع | 95.21 | 4.79 |
| ف | 96.12 | 3.88 |
| ك | 94.25 | 5.75 |
| م | 98.24 | 1.76 |
| Average | 95.07 | 4.93 |

To prove the ability of the proposed system to recognize Arabic handwritten characters, we tested 840 images out of three persons i.e. 3×28×10 that are collected, cropped, and segmented and the evaluation results are shown in Table V of the tested images.

### D. Experiment 2: Comparitive Evaluation

The comparative study of the proposed architecture is performed using the same database in [12]. Table VI shows the comparative evaluation of the proposed system compared with [12] as follows:

TABLE VI.    COMPARATIVE EVALUATION OF THE CRR AND ECR OF THE PROPOSED DCNN-SVM AND LOEY ET AL. [12]

| Author | CRR | ECR |
|---|---|---|
| Loey et al [12] | 94.90% | 5.10% |
| Proposed DCNN-SVM | 95.07% | 4.93% |

A rough comparison between the proposed architecture with the state-of-the-art is investigated in Table VII by which we compare the CRR, and ECR concerning the database used, and the trained, tested data.

TABLE VII.    COMPARATIVE EVALUATION OF THE PROPOSED DCNN-SVM WITH THE STATE OF THE ART SYSTEM

| Author | Database used | No. of Trained /Tested images | CCR (%) | ECR (%) |
|---|---|---|---|---|
| **Elleuch [9-10]** | HACDB images | 6600 images | 97.35% 24 class 93.41% 66 class | 5.83% 24 class, 2.09% 66 class |
| **Loey et al. [12]** | Arabic Handwritten Characters Dataset | 16800 images 13440 Training 3360 Testing images | 94.90% | 5.10% |
| **Proposed DCNN-SVM** | [12] and 840 additional images | 16800 images 13440 Training 3360 Testing images 840 tested images | 95.07% | 4.93% |

## V. CONCLUSION AND FUTURE WORK

This paper presents an effective deep convolutional neural network architecture used to extract and classify the Arabic handwritten characters. To ensure the reliability and efficiency of the proposed architecture, we apply a dropout support vector machine to classify and recognize the missing features that are not correctly classified by DCNN. Moreover, the proposed system divided the multi-stroke with similar Arabic characters to 13 clusters depends on K-means clustering. The proposed system achieves 95.07% correct classification accuracy with a minimum error classification rate of 4.93% compared with recent approaches. In the future, we plan to use full Arabic sentences that is maybe helpful for recognize and classify Arabic sentences as it considered one of the most challenges in the computer vision field.